\documentclass[letterpaper]{article}

\usepackage{stylefile}
\usepackage{times}
\usepackage{helvet}
\usepackage{courier}
\usepackage{url}
\usepackage[english]{babel} % For language and accents
\usepackage{latexsym}

\usepackage{adjustbox}
\usepackage{array}

\usepackage{multirow}
\usepackage{booktabs}

\newlength\tbspace
\newlength\rbspace
\newcolumntype{L}{c<{\hspace{\tbspace}}}
\newcolumntype{R}[2]{%
    >{\adjustbox{angle=#1,lap=\width-(#2)}\bgroup}%
    l%
    <{\egroup}%
}

\newcommand*\rot{\multicolumn{1}{R{38}{0.8em}}}

\usepackage{amsmath} % For math formulas
\usepackage{amssymb} % For \mathbb
\usepackage{bbm} % For \mathbbm

\usepackage{graphics}
\graphicspath{{./}}

\usepackage[colorinlistoftodos,prependcaption,textsize=small]{todonotes} 

\usepackage{titlesec}
\titleformat*{\paragraph}{\normalfont\bfseries\itshape}

\title{Learning Multi-Modal Word Representation Grounded in Visual Context}
\author{\'Eloi Zablocki \and Benjamin Piwowarski \and Laure Soulier \and Patrick Gallinari\\
Sorbonne Universit\'{e}s,
UPMC Univ Paris 06, UMR 7606, CNRS, LIP6, F-75005, Paris, France\\
\texttt{\{eloi.zablocki, benjamin.piwowarski, laure.soulier, patrick.gallinari\}@lip6.fr}
}

\begin{document}

\maketitle

\begin{abstract}
    Representing the semantics of words is a long-standing problem for the natural language processing community. Most methods  compute word semantics given their textual context in large corpora. More recently, researchers attempted to integrate perceptual and visual features. Most of these works consider the visual appearance of objects to enhance word representations but they ignore the visual environment and context in which objects appear. We propose to unify text-based techniques with vision-based techniques by simultaneously leveraging textual and visual context to learn multimodal word embeddings. We explore various choices for what can serve as a visual context and present an end-to-end method to integrate visual context elements in a multimodal skip-gram model. We provide experiments and extensive analysis of the obtained results.

\end{abstract}

\section{Introduction}
      
Representing word semantics is a long-standing problem that conditions major applications such as automatic translation \cite{bahdanau2014neural}, sentiment analysis \cite{maas2011learning}, and text summarization \cite{rush2015neural}. Distributional Semantic Models (DSMs) leverage large text corpora under the \textit{Distributional Hypothesis} \cite{harris1954distributional}, a strong assumption which states that \textit{words that occur in similar contexts should have similar meanings}, to produce fixed-length vectorial representation for words based on their co-occurrences in text corpora.

% TODO camera ready
%To further improve the quality of word representation, leveraging multimodal information is crucial. Indeed, psychological studies have given pieces of evidence that the meaning of words is grounded in perception \cite{glenberg2002grounding,Barsalou2008GroundedCognition} and \citeauthor{Gordon2013a} \shortcite{Gordon2013a} report a bias between what is said in texts and what can be seen in images. These observations outline the complementary roles of images and texts and bring new perspectives to multimodal approaches bridging textual information with visual ones to improve natural language processing tasks \cite{Hill2014,Lazaridou2015b}. Besides, it is worth mentioning that this has become possible thanks to the exploitation of significant advances in computer vision offering efficient tools for semantic extraction in images \cite{krizhevsky2012imagenet,Xu2015b}.
To further improve the quality of word representation, leveraging multimodal information is crucial. Indeed, psychological studies have given pieces of evidence that the meaning of words is grounded in perception \cite{glenberg2002grounding,Barsalou2008GroundedCognition} and \cite{Gordon2013a} report a bias between what is said in texts and what can be seen in images. These observations outline the complementary roles of images and texts and bring new perspectives to multimodal approaches bridging textual information with visual ones to improve natural language processing tasks \cite{Hill2014,Lazaridou2015b}. Besides, it is worth mentioning that this has become possible thanks to the exploitation of significant advances in computer vision offering efficient tools for semantic extraction in images \cite{krizhevsky2012imagenet,Xu2015b}.

\begin{figure}
    \centering
    \def\svgwidth{\linewidth} 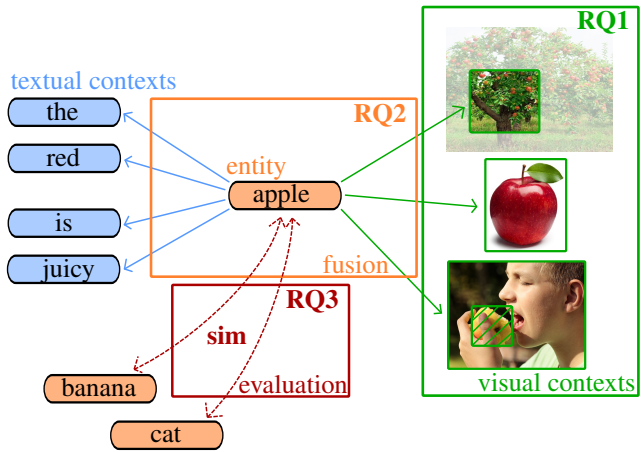
    \caption{\label{overview}Illustration of our approach and underlying research questions: RQ1 concerns the visual part of the model, RQ2 is about the integration of the visual part with the text model and RQ3 deals with the evaluation of the embeddings.}
\end{figure}
        
        \begin{figure*}[htpb]
	\centering
    \def\svgwidth{0.75\linewidth} 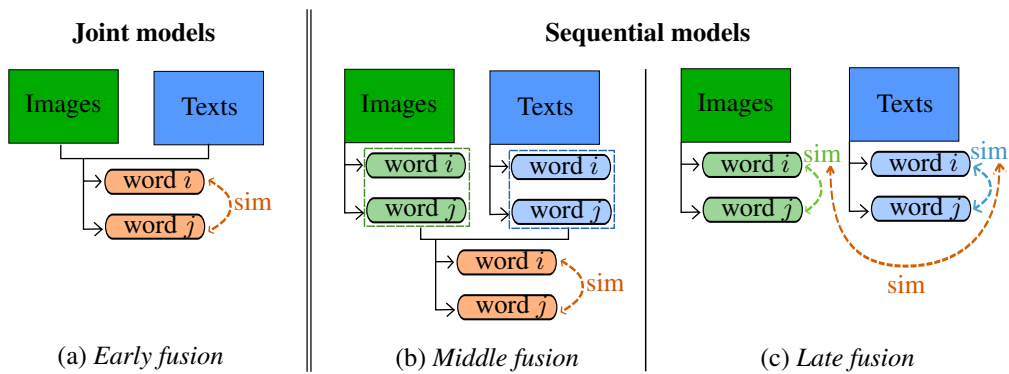
    \caption{\label{joint_sequential}Overview of early fusion, middle fusion, and late fusion techniques. Round-corner rectangles denote word embeddings. Green is related to images and blue to text, orange round-corner rectangles are multimodal embeddings built from textual and visual resources. \textquotedblleft sim" stands for an example of an evaluation task, namely \textit{word similarity}.}
\end{figure*}

In this context, multimodal representation learning models have been proposed to enhance word representations using either sequential \cite{Kiela2014,Bruni2014} or joint fusion techniques \cite{Hill2014,Lazaridou2015b}.
However, most of these works ignore the \emph{visual context} of objects. We posit that learning representations of contexts in  different modalities should be a key component of multimodal DSMs.
The importance of context is illustrated in a simple example (Figure~\ref{overview}). From an image of an apple on a black background, we can see its color, its texture and shape. From its context, e.g., growing on a tree, we can infer the relative size of apples with respect to the tree leaves, and that apples are fruits that grow on trees. If there is someone that is eating the apple, we can infer that apples are edible, and so on. From this example, we understand why exploiting the visual surroundings and context of objects might be  useful to grasp the semantics of words.

In this work, we propose a multimodal model for learning word representation, leveraging contexts in different modalities, namely texts and images.
Our contribution is threefold:
\begin{itemize}
	\item We propose and experiment with various definitions of what visual context is (Section \ref{rq1}) -- this has never been taken into account to the best of our knowledge in such models;
    \item We propose a multimodal context-driven model to jointly learn representations from textual and visual modalities, where both modalities influence media-independent word embeddings (Section \ref{rq2}). One further strength of the model is that it does not require aligned images and text (i.e. images with captions);
    \item We present a thorough analysis of the obtained results to determine the influence of the visual modality on the learned multimodal embeddings (Sections \ref{rq3} and \ref{rq32}) by experimenting with a set of word classification tasks.
\end{itemize}

\section{Related Work} 
\subsubsection{Learning word representation from textual resources.}
Distributional Semantic Models (DSMs) are implicitly or explicitly based on a factorization of a co-occurrence matrix to compute the representation of words. Well-known models are GloVe \cite{pennington2014glove} and  \textit{Word2Vec} \cite{Mikolov2013} on which we are based. In the latter, words are either predicted given their context (Continuous Bag Of Word model) or vice-versa (Skip-Gram model). In both cases, a representation is learned for both words and their context. 
Several modifications and improvements have been proposed to the Skip-Gram model, such as using Gaussian embeddings to account for the variance of the meaning of words \cite{vilnis2014word} and using extra information provided by Knowledge Bases \cite{tian2016learning}. 

\subsubsection{Learning word representations from textual and visual resources.}
% TODO camera ready
%Recent studies motivate the construction of general-purpose word embeddings with both language and perceptual inputs such as images. More precisely, psychological studies reveal that the meaning of words is grounded in perception \cite{glenberg2002grounding,Barsalou2008GroundedCognition}. Moreover, \citeauthor{Gordon2013a} \shortcite{Gordon2013a} highlight the complementarity of language and images. In particular, the \textit{Human Reporting Bias} states that the frequency with which people refer to things or actions in language does not correlate with real world frequencies. People usually do not mention common things, and  rather talk and write about surprising events. This systematic bias with respect to real-world frequencies motivates researchers to exploit visual information to learn word representation, leading to \textit{multimodal approaches}. 
Recent studies motivate the construction of general-purpose word embeddings with both language and perceptual inputs such as images. More precisely, psychological studies reveal that the meaning of words is grounded in perception \cite{glenberg2002grounding,Barsalou2008GroundedCognition}. Moreover, \cite{Gordon2013a} highlight the complementarity of language and images. In particular, the \textit{Human Reporting Bias} states that the frequency with which people refer to things or actions in language does not correlate with real world frequencies. People usually do not mention common things, and  rather talk and write about surprising events. This systematic bias with respect to real-world frequencies motivates researchers to exploit visual information to learn word representation, leading to \textit{multimodal approaches}. 

With this in mind, two main lines of multimodal DSM approaches have been proposed: sequential models and joint models, as illustrated in Figure~\ref{joint_sequential}. 

Sequential methods separately construct visual and textual word representations, and then combine them using different techniques, i.e. through \textit{middle fusion} or \textit{late fusion}. 
Given separately learned representations in each modality, middle fusion consists in merging them to form a multimodal vector (see Figure~\ref{joint_sequential} (b)). Several aggregation methods have been considered such as Concatenation \cite{Kiela2014b}, Singular Value Decomposition (SVD) \cite{Bruni2012}, Canonical Correlation Analysis (CCA) \cite{Silberer2012}, Weighted Gram Matrix Combination \cite{Hill2014e} or the task-driven cross-modal mapping \cite{Collell2017}. % TODO we can remove one element of the list here if too long
In late fusion (see Figure~\ref{joint_sequential} (c)), word representations are computed for each modality. Their multimodal interactions occur downstream in the task, as done in \cite{Bruni2014} who use a simple linear combination of similarity scores respectively obtained from textual and visual data.
In most of the sequential models cited above, textual representations are pre-trained Glove \cite{pennington2014glove} or Word2Vec \cite{Mikolov2013} embeddings and the visual embeddings are built from the aggregation (e.g.\ average or pooling) of activations obtained with a pre-trained CNN forwarded on images.

% Joint-models were proposed to directly learn a joint representation from textual and visual inputs (Figure \ref{joint_sequential}
% While middle and late fusion prevent potentially beneficial interactions due to having representations learned separately in each modality, joint-models...
While middle and late fusion prevent potentially beneficial interactions during training between the different modalities, 
joint models directly learn a joint representation from textual and visual inputs (Figure~\ref{joint_sequential} (a)). This idea is close to the way humans learn grounded meaning in semantics as observed in \cite{glenberg2002grounding} and \cite{Barsalou2008GroundedCognition}.
Some joint models require aligned texts and images. For example %\cite{Feng2010} and %TODO we can add this citation
\cite{Roller2013a} use a Bayesian modeling approach based on the assumption that text and associated images are generated using a shared set of underlying latent topics and \cite{Kottur2015b} ground word representations into vision by trying to predict the abstract scene associated to a given sentence. Our model follows an early fusion strategy but does not require aligned text and images.

Closer to our work, extensions of the \textit{Word2Vec} skip-gram were proposed. For example, \cite{Hill2014} base their model on the assumption that the frequency of appearance of concrete concepts correlates with the likelihood of \textquotedblleft experiencing" it in the world. Perceptual information for concrete concepts is then introduced to the model whenever that concept is encountered in the textual modality. Representations of concrete words are trained to predict surrounding words (as in the classical skip-gram model) and the perceptual features -- feature-norms defined in \cite{mcrae2005semantic} that describe objects as a set of features (typical color, usage, etc.). This work was later followed by \cite{Lazaridou2015b} whose method is designed to use natural images instead of the feature-norms which are constructed by hand. They force the representation of words for which they have images to be close to their visual (pre-trained) representation. Our work further exploit this line of research, but focuses on exploiting the visual context, which has not been done to the best of our knowledge.

\subsubsection{Using and modeling visual contexts.}
Several of the works presented above use the visual modality to constrain the textual representation to be close to the visual representation of the object. Such a strategy has two drawbacks. First, there is an asymmetry in the consideration of the modalities: text defines a semantic context for each word -- its surrounding words -- while images are used to have visual information about the object. Second, it does not use the fact that the context in which objects appear is informative and complementary to textual inputs to improve word representation. Indeed, this fact is supported by several works such as \cite{Bruni2012} who propose a middle fusion approach where a visual embedding is built by counting the number of visual words in images. This is the first attempt to apply the distributional hypothesis to images: \textit{Semantically similar objects will tend to occur in similar environments in images}. Through their experiments, they come to the conclusion that the appearance of the context (surrounding of objects) is more informative for semantics than the appearance of the object itself. In comparison to our model, their work does not propose to jointly learn embeddings from both visual and textual context.

This statement is strengthened with observations in \cite{Roller2013a} and \cite{Bruni2014}. The former proposes a Latent Dirichlet Allocation (LDA) model. The latter uses a count-based technique to learn multimodal word embedding by leveraging both visual and textual contexts. First, they build target-context count matrices for text (count of co-occurrence patterns with contexts) and images, using bag-of-visual words to represent images. They concatenate both matrices and perform rank reduction with SVD. They then split matrices (smoothed text and smoothed image matrices) and consider fusion at the feature level or scoring level. However, they use a \textquotedblleft count-base" method which does not learn representation for contexts and performs poorly on semantic tasks, moreover, their approach uses bags of visual words representation for images.

In addition to the identification of entities in their context, rich spatial information is present if objects can be located in the image. \cite{Bruni2014} propose to use this intrinsic spatial information for contexts by dividing the image in 4x4 bins and considering visual words separately for each region. However, when it comes to learning representations for words, exploiting spatiality is challenging and still largely under-explored.

\section{Research Questions}
From reviewing the literature, we observe three main issues with current multimodal DSM for which there are no consensus answers:
\begin{itemize}
	\item Text and images are very different by nature \cite{Gordon2013a}. A sentence has a linear structure with a list of tokens (words) while an image has spatially-organized quantifiable information (pixel values). In the skip-gram model, choosing surrounding words to be the context is a natural choice for a text, however, in images, it is not clear what should be used as context to learn semantically rich representations for objects \cite{Roller2013a,Bruni2014}).
    \item Several multimodal fusion methods exist, but none of the models presented above is significantly better  than the others, and the question to know how to build a multi-modal framework has no obvious answer, especially when the alignment between texts and images is missing.
    \item Evaluation tasks to assess the quality of word embeddings are inherently biased \cite{faruqui2016problems}, and it is hard to examine in depth the contribution brought by the visual modality \cite{Collell2016}.
\end{itemize}

In contrast to other works in learning multimodal word representations, we posit that exploiting the visual context enhances the learned representation of words. This assumption makes us consider images of complex scenes containing many objects. Indeed, images of a single object give very little information about the object, how it is used for, where it can be found and so on. On the contrary, an image showing an object in its environment, being used or interacting with other objects, is much more informative thanks to the surrounding context.
Accordingly, we address the following research questions, also illustrated in Figure~\ref{overview}:
    (\textbf{RQ1}) In images, what can be used to learn semantic representations for objects? In particular, does context can capture some of the semantic of a word/entity? Note that in this work, we consider that the set of entities is the subset of the set of words that correspond to objects in images.
    (\textbf{RQ2}) How can we naturally integrate a visual model with a text-based model to form a multimodal DSM?
    (\textbf{RQ3}) How can we evaluate and examine the contribution given by the visual modality in the final word embeddings?

\section{Model: Learning Multi-Modal Context-Driven Word Representations}

We present here a multimodal DSM model leveraging both visual and textual contexts of words in order to fulfill the distributional hypothesis.
To do so, we first formalize a definition of visual context and propose experiments to select appropriate visual context elements (RQ1). 

We then introduce our multimodal joint model based on the skip-gram framework  \cite{Mikolov2013} (RQ2). The textual and visual parts of the model share the same word embeddings which are updated from both textual and visual inputs, but contexts are modality specific. One strength of our model relies on the fact that it does not require aligned data.
Since this is not the focus of the paper, we assume that objects are already detected in images.

\subsection{Representation learning with visual contexts}
\label{rq1}
In this section, we formalize what we name \textit{visual contexts} and detail the choice of modeling that we propose. 

\subsubsection{Formalization.}
Based on the original Word2Vec skip-gram algorithm that considers entities $e$ (words) and their contexts $\mathcal{C}_e=\{c_1, ..., c_n\}$ ($n$ surrounding words within a window centered on the entity), we translate in what follows the distributional hypothesis for images to a concrete model.

In our case, the contexts are visual contexts that we define latter.
The choice for visual context elements $c \in \mathcal{C}_e$ does not need to correspond to a list of semantic entities \cite{levy2014}. For instance, visual context elements can be the surrounding objects, low-level features such as the visual appearance, or also the localization of the surrounding objects with respect to the considered entity. 

With this in mind, we define a function $f_\theta$, parametrized by $\theta$ (learned), such that for any entity $e$ and visual context element $c \in \mathcal{C}_e$, $f_\theta(c)$ is a vector of $\mathbb{R}^d$. These representations are then used in the negative-sampling loss:

\begin{align}
    \mathcal{L}_{\text{i}} & = - \sum\limits_{e \in \mathcal{D}} \sum\limits_{c \in \mathcal{C}_e} \biggl[ \log \sigma (f_\theta(c)^\top t_{e}) \notag \\
    & \qquad \qquad \qquad + \sum_{c^-} \log \sigma (- f_\theta(c^-)^\top t_{e})  \biggr]
\end{align}
where $\mathcal{D}$ is the set of entities, $t_e$ is the embedding associated to the entity/word $e$ (learned), $c^-$ is a negative context, and $\sigma$ is the sigmoid function.
This loss formulation is very close to the original skip-gram loss but integrates the learning of $f_\theta$ which shares parameters ($\theta$) for the computation of every context element. 

\subsubsection{Choice of modeling.}
Given an entity $e$, we propose different ways of modeling an instance of visual context elements $c \in \mathcal{C}_e$ and we detail how to build and parametrize $f_\theta$.

\paragraph{High-level context (surrounding objects). }
An image $I$ can be seen as a bag of objects: $I = \{o_1, o_2, ... \}$. This simple view gives high-level information about the environment in which objects occur. Given an entity $e = o_i$ (for some $i$) in an image, we define ${\mathcal{C}_e = \{o_j, j \neq i\}}$ as the set of all other objects that appear in the image. Then, a context ${c = o_j \in \mathcal{C}_e}$ is a surrounding object. We define ${f_\theta(c) = V_c}$  where ${V \in \mathbb{R}^{M \times d}}$ is a simple lookup table of embeddings for $M$ objects, $d$ the dimension of the representation space, and $V_c$ is the $c$\textsuperscript{th} row of this matrix.

\paragraph{Low-level context (image patches).}
At a coarser level, the set $\mathcal{C}_e$ of all visual context elements can be seen as image patches from the full image where entity $e$ is masked out with black pixels.
We call this \textit{low-level context} since it directly uses pixel values from the surroundings of entities. Using low-level context is interesting because some objects can be left unidentified in images by current models. However, this requires a bigger and more complex model and it is more difficult to extract meaningful information from pixel values. We suggest two possible choices to select $c \in \mathcal{C}_e$:
	(1) The instance $c$ is the full image where the entity is masked out by replacing RGB values with zeros;
    (2) $c$ is a small image patch randomly chosen around the entity. In practice, there are several choices for $c$ such that $c \in \mathcal{C}_e = \{c_1, c_2, ...\}$. 

In both cases, the image patch $c$ is forwarded in a CNN, parametrized by $\theta_1$, to form an activation vector ${u_c = \text{CNN}_{\theta_1}(c) \in \mathbb{R}^{B}}$ (where $B$ is the size of the last CNN filter, and equals 2048 in our experiments) obtained at the last layer of the network. The visual context vector ${f_\theta(c) = N u_c}$ is then formed with the projection of $u_c$ to the dimension $d$ with a matrix ${N \in \mathbb{R}^{d \times B}}$. Parameters to be learned are ${\theta = \{\theta_1, N\}}$.

\paragraph{Enhancing context with spatial information.}
When a dataset provides localization information for entities (i.e.\ bounding boxes or segmentation masks), we wish to use these annotations as it gives additional spatial information. For example, by looking at the position of a cup in an image with respect to a table or the hand of a person, one can infer that cups lie on tables and that they can be handed by people. We wish to enhance the visual contexts presented above with spatial information. We consider two methods to model what we name \emph{visual spatiality} to compute a vector $s_{(e, c)}$ representing the visual relationships between $e$ and $c$, and two models to integrate it with a visual context element $c$ as $f_\theta^{sp}(c, s_{(e, c)}) \in \mathbb{R}^d$.

The first method considers low-level features, and corresponds to a 4-d spatial vector whose components are the relative positions on the $x$ and $y$ axes of the two bounding boxes of $e$ and $c$ (denoted $\delta_x$ and $\delta_y$), and the ratio of width and height between the two bounding boxes of $e$ and $c$ ($\delta_\text{width}$ and $\delta_\text{height}$). The second method is a high-level features vector, and corresponds to a 4-d spatial vector whose components are four indicator functions denoting whether the context $c$ is below, beside, above, or bigger than the entity $e$ ($1$ if true, $0$ otherwise). Following \cite{Ludwig2016}, the context is said to be \textquotedblleft below" its entity if $|\delta_x| \leq \delta_y$, \textquotedblleft above" its entity if $|\delta_x| \leq -\delta_y$ and \textquotedblleft beside" otherwise. A context is said to be bigger than its entity if $\delta_\text{width} \delta_\text{height} \geq 1$.

Once the spatial vector $s_{(e, c)}$ is built, it is integrated with the visual context embedding ${v_c = f_\theta(c)}$, to form a spatially-informed visual context ${v^{sp}_c = f_\theta^{sp}(c, s_{(e,c)})}$ that is used in the skip-gram equations instead of $f_\theta(c)$. Again, two variants are considered: (1) a linear combination of the visual context $v_c$  with the spatial vector $s_{(e, v)}$, i.e. $f_\theta^{sp}(c, s_{(e, c)}) = M.(v_c \oplus s_{(e, c)})$ where $M \in \mathbb{R}^{d \times (d+4)}$ and $\oplus$ denotes the concatenation operator; (2) a bilinear interaction ${f_\theta^{sp}(c, s_{(e, c)}) = s_{(e, c)} M v_c}$ where $M \in \mathbb{R}^{4 \times d \times d}$. This model has more free parameters but considers a bilinear interaction between the spatial vector $s_{(e, c)}$ and the visual context $v_c$.

\begin{table*}[ht!]
	\centering
	\setlength{\tabcolsep}{5.6pt} % Space between columns 6pt is standard 
	\begin{tabular}{c | c | r ||@{\hspace{\rbspace}} c  c  c  c  c<{\hspace{\rbspace}} ||@{\hspace{\rbspace}} c  c  c  c  c  c  c  c  c<{\hspace{\rbspace}}}
    	\multicolumn{3}{c@{\hspace{\rbspace}}||@{\hspace{\rbspace}}}{} & \rot{VisSim} & \rot{SemSim} & \rot{Simlex} & \rot{MEN} & \rot{WordSim} & \rot{Encyclopedic} & \rot{Taste} & \rot{Sound} & \rot{Taxonomic} & \rot{Function} & \rot{Tactile} & \rot{Color} & \rot{Shape} & \rot{Motion} \\
        \hline
    	\multicolumn{3}{c@{\hspace{\rbspace}}||@{\hspace{\rbspace}}}{} & \multicolumn{5}{c@{\hspace{\rbspace}}||@{\hspace{\rbspace}}}{Similarity Evaluation} & \multicolumn{9}{c@{\hspace{\rbspace}}}{Feature-norm Prediction Task}\\
        \hline
        \hline
        \multicolumn{2}{c}{baseline} & \textbf{L} & 43 & 45 & 16 & 22 & 17 & 56 & 49 & 36 & \textbf{76} & \textbf{56} & \textbf{17} & \textbf{41} & \textbf{60} & \textbf{58}\\
        \hline
    	\multirow{8}{*}{\rotatebox[origin=c]{90}{\textbf{Our models}}} & Objects & \textbf{O} & 43 & 54 & 31 & 64 & 27 & 48 & 46 & 35 & 62 & 48 & 03 & 21 & 43 & 36\\
        \cline{2-17}
    	& \multirow{2}{*}{Patches} & \textbf{P} & 28 & 35 & 17 & 35 & 22 & 30 & 51 & 23 & 48 & 37 & 04 & 24 & 38 & 30\\
    	& & $\textbf{P}_\text{full}$ & 35 & 42 & 19 & 43 & 28 & 30 & 48 & 30 & 46 & 35 & 06 & 23 & 35 & 27\\
        \cline{2-17}
        & \multirow{4}{*}{Spatial} & $\textbf{Sp}(\textbf{O}, \delta, \oplus)$ & 48 & 57 & 32 & 58 & 27 & 40 & 55 & 28 & 54 & 50 & 06 & 24 & 44 & 37\\
        & & $\textbf{Sp}(\textbf{O}, c, \oplus)$ & 48 & 58 & 30 & 58 & 25 & 40 & \textbf{60} & 33 & 54 & 50 & 11 & 25 & 41 & 34\\
        & & $\textbf{Sp}(\textbf{O}, \delta, b)$ & 46 & 56 & \textbf{35} & 54 & 28 & 37 & 57 & 27 & 50 & 50 & 15 & 24 & 38 & 32 \\
        & & $\textbf{Sp}(\textbf{O}, c, b)$ & \textbf{51} & \textbf{61} & 33 & 62 & 30 & 38 & 58 & 27 & 58 & 47 & 10 & 22 & 43 & 34\\
        \cline{2-17}
        & Ensemble & $\textbf{L} + \textbf{O}$ & 45 & 57 & 33 & \textbf{66} & \textbf{34} & \textbf{58} & 52 & \textbf{42} & 74 & \textbf{56} & 02 & 27 & 53 & 53\\
    \end{tabular}
    \caption{\label{results1} RQ1 results. The columns on the left part of the table are the Spearman correlations (multiplied by 100) on the word similarity benchmarks (only word pairs with visual entities are evaluated). The columns on the right side are the f1-scores (multiplied by 100) at the feature-norm prediction task (grouped by feature category as proposed in \cite{Collell2016}). Best results are highlighted in bold.}
\end{table*}

\subsection{Integration in a multimodal model}
\label{rq2}
\label{integration_multi}
We now present our multimodal representation learning model that integrates the previously presented visual module with the textual skip-gram. 
The main idea is that while word embeddings should be shared across modalities, context is media-specific. 
The contribution of each modality is controlled by a linear combination (hyper-parameter $\alpha$, determined by cross-validation) of modality-specific costs, which gives the following global loss function:

\begin{equation}
\label{eq:multi}
\mathcal{L}(T, U, \theta) = \mathcal{L_\textnormal{t}} (T, U) + \alpha \mathcal{L_\textnormal{i}} (T, \theta)
\end{equation}
where $U$ denotes the textual context lookup table and $\mathcal{L_\textnormal{t}} (T, U)$ is the \textit{Word2Vec} loss function \cite{Mikolov2013}.

A crucial point is that this model does not require aligned texts and images to train the model, or extra pre-trained representations on external datasets -- we only require that entities identified in images to be associated with a unique word of the vocabulary. Besides, we justify the use of a joint model as we think it is important that representations are learned both for entities and for contexts. Indeed, as the entities embeddings are affected by both modalities, the context representations should change and be updated by transitivity between modalities through the shared embeddings.

\section{Evaluation protocol}
\label{rq3}
In this section, we evaluate word embeddings on different tasks. In particular, we measure the performance of word embeddings built from visual data (RQ1) and multimodal data (RQ2).
 
\subsection{Data}
We use a large collection of English texts, a dump of the Wikipedia database ({http://dumps.wikimedia.org/enwiki}), cleaned and tokenized with the Gensim software \cite{rehurek_lrec}. This provides us with $4.2$ million articles, and a vocabulary of $2.1$ million unique words.
For visual data, we use the Visual Genome dataset \cite{krishnavisualgenome} as it is a large image collection (108k images) with a large number of different objects (4842 unique entities with more than 10 occurrences) in rich and complex scenes (31 object instances per image on average).

\subsection{Scenarios and Baselines}

\subsubsection{Scenarios.}
To evaluate the different components of our model, we evaluate different scenarios. In particular, we train the model that uses other objects as visual contexts (noted \textbf{O}), the model that uses image patches (\textbf{P}) and the model that uses full images ($\textbf{P}_\text{full}$).

Models that use spatial context information are also evaluated and are denoted $\textbf{Sp}(.,.,.)$ where 
the first argument denotes the visual context type (\textbf{O}, \textbf{P} or $\textbf{P}_\text{full}$), the second the spatial context features ($\delta$ or $c$), and the third the integration ($\oplus$ for concatenation and $b$ for bilinear product integration). For instance, $\textbf{Sp}(\textbf{P},\delta,b)$ corresponds to using image patches, with low-level visual features and bilinear product.

All combinations of those models with the skip-gram text-only model (\textbf{T}) are trained and evaluated to get multimodal word representations, with the method explained in section \ref{integration_multi}.

\subsubsection{Baselines}

Our baseline (\textbf{L}) is inspired by the state-of-the-art model of \cite{Lazaridou2015b}, since they use visual features from objects themselves to learn word representations in contrast to the visual context features we use in our model.
% However, we modified the visual part loss to use higher level features (objects) to analyze the true difference between using context and visual aspect.
For any visual entity $e$, they assume that a visual vector $v_e$ representing the entity is available. During training, along with the text-only skip-gram loss, the similarity between the embedding of the entity and its visual appearance is maximized in a max-margin framework:
\[ \mathcal{L}_{\text{object}} = \sum\limits_{e \in \mathcal{D}} \sum\limits_{v^{-}} \max (0, \gamma - \cos (t_e, v_e) +  \cos (t_e, v^{-})) \]
where $\gamma$ is the margin and $v^{-}$ is the visual appearance of a ``negative'' object (random). For an object $e$, $v_e$ is kept fixed and visual information is incorporated each time the entity is encountered in text.
We note this model $\textbf{L} + \textbf{T}$ where \textbf{L} corresponds to the visual loss and \textbf{T} the text-only skip-gram loss.

To evaluate our visual context-driven multimodal representation learning model (RQ2), we also evaluate: 1) the skip-gram text only model (noted \textbf{T}), and 2) a sequential model, noted $\textbf{O} \oplus \textbf{T}$, where embeddings of model $\textbf{T}$ are concatenated with embeddings obtained from $\textbf{O}$ and then projected in a lower-dimensional space with PCA. This serves as a comparison point between our joint approach and a sequential one.
%and its combination with the baseline model ($\textbf{B} + \textbf{T}$) where the sign '$+$' denotes the summation of the loss function of the two models when the embeddings for entities are shared.

\subsection{Tasks}
Similarly to previous work \cite{Lazaridou2015b,Collell2017}, we evaluate our model on three different semantic tasks, namely word similarity and relatedness, feature norm prediction, and abstractness/concreteness prediction. Each task serves as a biased indicator of the quality of the embeddings.
 We present these evaluation benchmarks in what follows.
\subsubsection{Word similarity and relatedness benchmarks.}
Semantic relatedness (resp.\ similarity) evaluates the similarity (resp.\ relatedness) degree of word pairs. We use several benchmarks which provide gold labels (i.e.\ human judgment scores) for word pairs: WordSim353 \cite{Finkelstein2002}, MEN \cite{Bruni2014}, SimLex-999 \cite{Hill2015}, SemSim and VisSim \cite{Silberer2014a}. The spearman correlation is computed between the list of similarity scores given by the model (cosine-similarity between multimodal vectors) and the gold labels. The higher the correlation is, the more semantic is captured in the embeddings. While word similarity benchmarks are widely used for intrinsic embedding evaluation, they are biased in the sense that good intrinsic evaluation scores do not imply useful embeddings for downstream tasks as shown by \cite{faruqui2016problems}.

\subsubsection{Feature norm prediction.}
% TODO camera ready
%\citeauthor{Collell2016} \shortcite{{Collell2016}} use the task of predicting features norms (e.g.\ "is\_red", "can\_fly") of objects given word representation to evaluate visual or textual-based representations.
\cite{Collell2016} use the task of predicting features norms (e.g.\ `is\_red', `can\_fly') of objects given word representation to evaluate visual or textual-based representations.
%In particular, they compare how the different representations learned from either modality help for the prediction of feature norms of entities.
%Beside, \cite{Bulat2016} consider the zero-shot learning task to predict properties for images of unseen concepts, thus learning a cross-modal map between a multi-modal (text + image) space to the property-norm space. To build the multi-modal space, they concatenate Word2Vec features of a concept along with the average pre-soft activation of the top-n images given by Google images on a pre-trained CNN. Then they learn the mapping between the multi-modal space to the feature-norm space by regression.
We consider this task to evaluate our word embeddings and use the same setup for evaluation. The evaluation dataset is an extract of the McRae dataset \cite{mcrae2005semantic}. There is a total of 43 characteristics grouped into 9 categories for 417 entities. A linear SVM classifier is trained and 5-fold validation scores are reported.

\begin{table*}[ht!]
	\centering
	\setlength{\tabcolsep}{5pt} % Space between columns 6pt is standard 
	\begin{tabular}{c | c | r ||@{\hspace{\tbspace}} c c c c L ||@{\hspace{\tbspace}} c c c c c c c c L || c}
    	\multicolumn{3}{c@{\hspace{\tbspace}}||@{\hspace{\tbspace}}}{} & \rot{VisSim} & \rot{SemSim} & \rot{Simlex} & \rot{MEN} & \rot{WordSim} & \rot{Encyclopedic} & \rot{Taste} & \rot{Sound} & \rot{Taxonomic} & \rot{Function} & \rot{Tactile} & \rot{Color} & \rot{Shape} & \rot{Motion} & \\
        \hline
    	\multicolumn{3}{c@{\hspace{\tbspace}}||@{\hspace{\tbspace}}}{} & \multicolumn{5}{c@{\hspace{\tbspace}}||@{\hspace{\tbspace}}}{Similarity Evaluation} & \multicolumn{9}{c@{\hspace{\tbspace}}||}{Feature-norm Prediction Task} & \multicolumn{1}{c}{Conc.} \\
        \hline
        \hline
        \multirow{3}{*}{\rotatebox[origin=c]{90}{\textbf{Basel.}}} & Text & \textbf{T} & 48 & 60 & 33 & 69 & 63 & 58 & 52 & 44 & 79 & 62 & 11 & 32 & 54 & 60 & 42.1\\
        \cline{2-18}
        & Sequential & $\textbf{O} \oplus \textbf{T}$ & 49 & 62 & 33 & 71 & 64 & 63 & 55 & 40 & 72 & 59 & 12 & 35 & 54 & 58 & 43.7\\
        \cline{2-18}
        & Joint & $\textbf{L} + \textbf{T}$ & 52 & 65 & 34 & 71 & 65 & 61 & 55 & 42 & 80 & 59 & 11 & 31 & 54 & 62 & 43.4\\ %bash /net/big/zablocki/tensorboard.sh /net/naivenewbeaters/zablocki/important_models/2017-07-06_17-31-35_text_laz/ 6022 naivenewbeaters
        \hline
        \multirow{8}{*}{\rotatebox[origin=c]{90}{\textbf{Our models}}} & Objects & $\textbf{O} + \textbf{T}$ & 53 & 66 & 35 & \textbf{75} & \textbf{67} & 62 & 55 & 46 & \textbf{82} & 61 & 13 & 33 & 55 & 61 & 42.9\\ %bash /net/big/zablocki/tensorboard.sh é" 6042 pascal
        \cline{2-18}
    	& \multirow{2}{*}{Patches} & $\textbf{P} + \textbf{T}$ & 53 & 65 & 35 & 72 & \textbf{67} & 60 & 56 & 49 & \textbf{82} & 60 & 12 & 32 & 55 & 61 & 43.1\\
    	& & $\textbf{P}_\text{full} + \textbf{T}$ & 53 & 65 & 34 & 73 & 65 & 60 & 55 & 44 & \textbf{82} & \textbf{63} & 14 & 32 & 55 & 59 & 43.2\\%bash /net/big/zablocki/tensorboard.sh /net/pascal/zablocki/important_models/2017-06-30_17-55-58_text_image_/ 6018 pascal
        \cline{2-18}
        & \multirow{4}{*}{Spatial} & $\textbf{Sp}(\textbf{O}, \delta, \oplus) + \textbf{T}$ & 52 & 66 & 36 & 73 & 64 & \textbf{64} & \textbf{59} & 46 & 81 & 62 & 06 & 31 & \textbf{57} & \textbf{63} & 42.5\\ % bash /net/big/zablocki/tensorboard.sh /net/pas/zablocki/important_models/2017-07-05_16-52-06_text_sem/ 6043 pas
        & & $\textbf{Sp}(\textbf{O}, c, \oplus) + \textbf{T}$ & 54 & 66 & 35 & 72 & 64 & 62 & 56 & \textbf{52} & 80 & 61 & 13 & \textbf{34} & \textbf{57} & 58 & 43.7 \\ %bash /net/big/zablocki/tensorboard.sh /net/titan/zablocki/saved_models/2017-07-28_11-25-45_text_sem_/ 6029 titan
        & & $\textbf{Sp}(\textbf{O}, \delta, b) + \textbf{T}$ & 54 & \textbf{68} & \textbf{38} & 73 & 66 & 63 & 56 & 48 & 81 & 60 & 13 & 32 & 56 & \textbf{63} & 42.5\\ %bash /net/big/zablocki/tensorboard.sh /net/pas/zablocki/important_models/2017-07-10_13-39-25_text_sem/ 6046 pas
        & & $\textbf{Sp}(\textbf{O}, c, b) + \textbf{T}$ & \textbf{55} & 67 & 34 & \textbf{75} & 64 & 61 & 58 & 46 & 80 & \textbf{63} & \textbf{15} & \textbf{34} & \textbf{57} & 62  & \textbf{44.4}\\%bash /net/big/zablocki/tensorboard.sh /net/pas/zablocki/important_models/2017-07-10_13-18-51_text_sem/ 6045 pas -- bash /net/big/zablocki/tensorboard.sh /net/titan/zablocki/saved_models/2017-07-31_17-42-20_text_sem_/ 6051 titan
        \cline{2-18}
        & Ensemble & $\textbf{L} + \textbf{O} + \textbf{T}$ & 54 & 66 & 35 & \textbf{75} & 65 & 63 & 55 & 50 & \textbf{82} & 60 & 10 & 33 & 55 & 59 & 43.9\\
    \end{tabular}
    \caption{\label{results2}RQ2 experimental results on word similarity evaluation benchmarks, feature-norm prediction task, concreteness prediction task (Conc.). Concreteness measures are coefficients of determination ($R^2$) multiplied by 100.}
\end{table*}

\subsubsection{Abstractness / Concreteness prediction.}
The USF norms \cite{nelson2004university} give concreteness ratings for 3260 English words. With a multimodal word representation, we wish to know if it contains information that can be used to predict the concreteness rating of the associated word. In practice, we train an SVM with a RBF kernel to predict the gold concreteness rating from word embeddings.
Note that this task is only used to evaluate multimodal representations since visual-based ones cover too small a vocabulary.

\subsection{Implementation details}
Experiments use python and Tensorflow \cite{abadi2016tensorflow}.
Images are upscaled to the shape $598 \times 598$ and passed through a pre-trained Inception-V3 CNN \cite{szegedy2016rethinking} to give spatial visual tensor of shape $17 \times 17 \times 2048$ (before the ReLU at the \textquotedblleft Mixed\_7c" layer). One slice of the tensor with a shape $1 \times 1 \times 2048$ corresponds to the activation of a region of the original image.
We use 5 negative examples per entity, and our models are trained with stochastic gradient descent with learning rate $l_r=10^{-3}$ and mini-batches of size 64. $N$ and $M$ are regularized with a $L_2$-penalty respectively weighted by scalars $\lambda$ and $\mu$. The values of hyperparameters were found with cross-validation: ${\lambda = 0.1}$, ${\mu = 0.1}$, ${\gamma = 0.5}$, ${\alpha = 0.2}$.

\section{Experiments and Results}
\label{rq32}

% Answering RQ3 enables us to gain insight about what is learned with the visual modality in comparison to traditional text-only approaches. To do so, we propose various complementary experiments: measuring the prediction accuracy of the concreteness and abstractness of words given their representations and predicting characteristics of objects given their embeddings (feature-norm prediction task). 

\paragraph{RQ1: Evaluating visual context-driven semantic representations of words.}
Table \ref{results1} reports the results of the experiments for RQ1 discussing what kind of visual information can be useful.

The first conclusion we draw is that surroundings of entities are more informative than the visual appearance of objects for the evaluation on all of the word similarity benchmarks. Indeed, results of the word similarity task highlight that our model scenarios generally overpass baselines. For instance,  results of our model $\textbf{P}_{\text{full}}$ is on average 29\% higher  than those of the baseline \textbf{L}.  However, on the feature-norm prediction task, direct visual features from objects (model \textbf{L}) are better suited for the categories that describe visually the objects (e.g.\ \textit{is\_red} in `Color' category or \textit{is\_round} in the `Shape' category) but not for the other non visual categories such as `Encyclopedic', `Taste' and `Sound'. 

To measure the complementarity of the features from objects and from their surroundings, we also evaluated an ensemble model that combines the baseline \textbf{L} and the \textbf{O} model ($\textbf{L} + \textbf{O}$) where '$+$' denotes the summation of the loss functions when the embeddings are shared. 
Interestingly, combining visual contexts and direct features (${\textbf{L} + \textbf{O}}$) results in a model that has a very good average performance, showing the complementarity of visual contexts with visual entity representations.

Our second observation shows that using spatial information is useful: performance is better on the word similarity benchmarks (+9\% improvement on average for $\textbf{Sp}(\textbf{O}, c, b)$ w.r.t.\ \textbf{O}) and the feature-norm prediction task (+20\%). Both high and low-level spatial features lead similar results. 
This reinforces our intuition that visual context, and more particularly spatial information, are promising for learning word representation and reducing the \textit{Human Reporting Bias} affecting texts and images.

The third conclusion we draw is that high-level contexts (in \textbf{O}) yield better scores  (+31\%) than low-level contexts (\textbf{P} or $\textbf{P}_\text{full}$). 
Using low-level visual features is a challenging problem. However, they are promising since they are cheap to collect, do not require context annotations, and contain rich information if handled correctly. The difficulty lies in the natural noise in the surroundings of objects and the need for visual modules that automatically extract high-level information from raw pixel values.

\paragraph{RQ2/RQ3: Evaluating our multimodal context-driven multimodal representation learning model / analysis.} 

Table \ref{results2} reports the results on RQ2 and RQ3. Embeddings are initialized with pre-trained embeddings obtained from the text-only baseline.

Results highlight that all of the trained multimodal outperform the text-only baseline for all evaluation tasks. For instance, $\textbf{O} + \textbf{T}$ shows an average improvement of 9\%  over \textbf{T}. This is in-line with the conclusions of related works \cite{Hill2014e}. Besides, a joint model (e.g.\ $\textbf{O} + \textbf{T}$) compares favorably to a sequential model ($\textbf{O} \oplus \textbf{T}$) built from embeddings obtained from $\textbf{O}$ and $\textbf{T}$ as we note a 5\% relative improvement, showing that embeddings computed using multiple modalities at once are beneficial. Like we did for RQ1, we also evaluated an ensemble model ($\textbf{L} + \textbf{O} + \textbf{T}$) to measure the complementarity of visual features in the multimodal model. Again, we generally notice a slight improvement over both $\textbf{O} + \textbf{T}$ and $\textbf{L} + \textbf{T}$. This opens perspectives for formalizing and leveraging visual information from both entities and their context.

The obtained results are consistent with the conclusions drawn above on the RQ1 analysis: visual surroundings of entities are more useful than direct features on the evaluated tasks (3.2\% improvement); the combination of both models shows the complementarity of the approaches, adding a spatial term for visual context significantly increases performances (6\% improv.); finally, higher-level contexts are slightly easier to use than lower-level contexts (1\% improv.).

To get a deeper insight into learned embeddings, we aim at explaining the impact of the visual modality on the multimodal word representation. To do so, we estimate the correlation between the shift measured on the embedding and the concreteness degree of a word. The result outlines a correlation of $\rho_\text{Spearman} = 0.33$, showing that visual and concrete words see their embeddings being more changed than other non visual and abstract words. This was to be expected because the visual part only adds information to visual entities.

%Finally, we looked at the learned embeddings. We give here a brief overview of the main findings.
%Visual and concrete words see their embeddings being changed more than other non visual and abstract words. This was to be expected because the visual part only adds information to visual entities. This statement is reinforced by a correlation between the shift measured on the embedding and the concreteness degree of a word ($\rho_\text{Spearman} = 0.33$).

\section{Conclusion and Future Work}
In this work, we proposed a multimodal (text and image) context-based approach to learn word embeddings. Through extensive experiments,  and in line with related work, we observed the complementarity of visual and textual data to learn word representations. More importantly, we have shown that visual surroundings of objects and their relative localization are very informative to build word representations -- actually, more than, but complementary to, the visual appearance of the objects themselves as exploited in previous works.

In future work, we will explore the use of downstream tasks to evaluate multimodal word embeddings as it might give finer insights on the way the visual part of the model contributes to learning representations.
Orthogonally, we will focus on contexts and their learned representations. In particular, we would like to see if aligned and consistent multimodal representations are learned with weak supervision provided by the entities.
Also, we will extend our work to learn relation representations between objects based on multimodal representations and the exploitation of existing knowledge bases.

%TODO camera ready
% TODO remove footnote
\section*{Acknowledgments}
This work is partially supported by the CHIST-ERA EU project MUSTER \footnote{http://www.chistera.eu/projects/muster} (ANR-15-CHR2-0005) and the Labex SMART (ANR-11-LABX-65) supported by French state funds managed by the ANR within the Investissements d'Avenir programme under reference ANR-11-IDEX-0004-02. We additionally thank Guillem Collell for providing pre-trained visual vectors needed for evaluating the baseline.

\fontsize{9.5pt}{10.5pt}
\selectfont
\bibliography{biblio.bib}
% TODO camera-ready 
%\bibliographystyle{aaai}
\bibliographystyle{apalike}
\end{document}